\documentclass[10pt,twocolumn,letterpaper]{article}
\usepackage{multirow}
\usepackage{float}
\usepackage[pagenumbers]{cvpr} 

\usepackage{graphicx}
\usepackage{svg}

\definecolor{cvprblue}{rgb}{0.21,0.49,0.74}
\usepackage[pagebackref,breaklinks,colorlinks,allcolors=cvprblue]{hyperref}

\def\paperID{*****} 
\def\confName{CVPR}
\def\confYear{2026}

\title{Multimodal Emotion Recognition via Bi-directional Cross-Attention and Temporal Modeling}

\author{Junhyeong Byeon \qquad Jeongyeol Kim \qquad Sejoon Lim\thanks{Corresponding author.}\\
Kookmin University, Republic of Korea\\
{\tt\small \{junhyeong0519, kjy013125, lim\}@kookmin.ac.kr}
}
\begin{document}
\maketitle

\begin{abstract}
Expression recognition in in-the-wild video data remains challenging due to substantial variations in facial appearance, background conditions, audio noise, and the inherently dynamic nature of human affect. Relying on a single modality, such as facial expressions or speech, is often insufficient for capturing these complex emotional cues. To address this limitation, we propose a multimodal emotion recognition framework for the Expression (EXPR) task in the 10th Affective Behavior Analysis in-the-wild (ABAW) Challenge.

Our framework builds on large-scale pre-trained models for visual and audio representation learning and integrates them in a unified multimodal architecture. To better capture temporal patterns in facial expression sequences, we incorporate temporal visual modeling over video windows. We further introduce a bi-directional cross-attention fusion module that enables visual and audio features to interact in a symmetric manner, facilitating cross-modal contextualization and complementary emotion understanding. In addition, we employ a text-guided contrastive objective to encourage semantically meaningful visual representations through alignment with emotion-related text prompts.

Experimental results on the ABAW 10th EXPR benchmark demonstrate the effectiveness of the proposed framework, achieving a Macro F1 score of 0.32 compared to the baseline score of 0.25, and highlight the benefit of combining temporal visual modeling, audio representation learning, and cross-modal fusion for robust emotion recognition in unconstrained real-world environments.
\end{abstract}    
\section{Introduction}
\label{sec:intro}

Emotion recognition from in-the-wild videos is a fundamental problem in affective computing, with broad applications in healthcare, education, and affect-aware interactive systems. However, robust recognition in real-world settings remains challenging because affective cues are often degraded by head pose variation, illumination changes, occlusion, background noise, and other uncontrolled factors~\cite{zafeiriou2017aff,kollias2019deep,kollias2021affect,kollias2024behaviour4all}.

To address these challenges, recent studies have increasingly explored multimodal frameworks that jointly leverage visual and audio signals~\cite{ngiam2011multimodal,tsai2019multimodal,tran2022pre,venkatraman2024multimodal,zhang2024affective}. The motivation is that the two modalities provide complementary affective cues: facial expressions capture appearance-based emotional information, while speech contains prosodic and paralinguistic patterns that can help resolve visually ambiguous cases. In addition, large-scale pre-trained models have demonstrated strong representation power in both visual and audio domains, making them promising foundations for multimodal emotion recognition~\cite{radford2021learning,baevski2020wav2vec20frameworkselfsupervised}.

Despite this progress, several challenges remain. First, many multimodal approaches do not fully exploit mutual interactions between visual and audio streams, limiting their ability to capture complementary cross-modal context~\cite{tsai2019multimodal,tran2022pre,venkatraman2024multimodal,zhang2024affective}. Second, emotional expressions evolve over time, and frame-level features alone are often insufficient for modeling the temporal dynamics required for robust prediction~\cite{bai2018empirical,lea2016temporal,zhao2021former,ma2022spatio,li2022nrdfernet,zhou2025emotion}.

In this paper, we propose a multimodal framework for the Expression (EXPR) Recognition track of the 10th Affective Behavior Analysis in-the-wild (ABAW) Challenge~\cite{kollias2025emotions}. Our method builds on frozen CLIP~\cite{radford2021learning} and Wav2Vec 2.0~\cite{schneider2019wav2vec,baevski2020wav2vec20frameworkselfsupervised} backbones for visual and audio feature extraction, respectively, and improves their integration in three ways. First, we apply a Temporal Convolutional Network (TCN) to frame-level visual features to model short-term temporal dependencies in facial expression sequences~\cite{lea2016temporal,bai2018empirical}. Second, we introduce a bi-directional cross-attention fusion module that allows visual and audio representations to refine each other through symmetric cross-modal interaction. Third, we regularize the pooled visual representation with CLIP text embeddings via a text-guided contrastive objective, encouraging semantically consistent representations for expression recognition~\cite{radford2021learning,zhou2022conditional,zhou2025emotion}.

The main contributions of this work are summarized as follows:
\\
\begin{enumerate}
    \item \textbf{Temporal visual modeling with TCN:} We enhance frame-level visual features with a Temporal Convolutional Network to capture temporal dependencies in facial expression sequences.
    \item \textbf{Bi-directional cross-attention fusion:} We introduce a symmetric cross-attention mechanism that enables bidirectional interaction between visual and audio features for more effective multimodal integration.
    \item \textbf{Label-based prompt supervision:} We convert expression labels into natural language prompts and use the resulting CLIP text embeddings as an auxiliary supervision signal for expression recognition.
\end{enumerate}
\section{Related Work}

\subsection{Facial Expression Recognition}
Facial Expression Recognition (FER) is a core task in affective computing that aims to identify human emotional states from facial images and videos. Recent studies have shown that deep learning-based visual representations are effective for recognizing discrete facial expressions~\cite{zafeiriou2017aff,kollias2019deep,kollias2019face}.

As FER research has progressed, increasing attention has been given to affective behavior analysis in unconstrained real-world settings. Accordingly, recent studies have focused on robust visual representation learning, temporal modeling for dynamic expressions, and multimodal fusion strategies for improved recognition performance. To address these issues, large-scale in-the-wild datasets such as Aff-Wild and Aff-Wild2 were introduced, providing more realistic benchmarks for facial behavior analysis~\cite{zafeiriou2017aff,kollias2019deep,kollias2019expression}. These datasets also supported studies on multiple affective tasks including expression recognition, valence-arousal estimation, and action unit detection~\cite{kollias2019face,kollias2021affect}.

Building upon these datasets, the ABAW challenge series has evolved over the years and has played an important role in advancing affective behavior analysis under realistic in-the-wild conditions~\cite{kollias2020analysing,kollias2021analysing,kollias2022abaw,kollias2023abaw2,kollias20246th,kollias20247th,kollias2025advancements,kollias2025emotions}. Related datasets, toolkits, and benchmarks have also been introduced to support broader behavior analysis tasks in real-world scenarios~\cite{kollias2025dvd,kollias2024behaviour4all}. In particular, the EXPR task in ABAW focuses on classifying facial expressions under challenging in-the-wild conditions, making it a representative benchmark for robust expression recognition. Recent ABAW studies have explored various learning strategies, including multi-task learning, compound expression recognition, and distribution-aware optimization, to improve performance on expression recognition and related affective tasks~\cite{kollias2023multi,kollias2023abaw,kollias2021distribution,kollias2024distribution}.

Despite these advances, facial expression recognition in the wild remains challenging due to subtle facial variations, ambiguous expression boundaries, and severe environmental noise.

\subsection{Vision-Language Emotion Recognition}

Recent advances in vision-language representation learning have enabled models to align visual content with semantic information expressed in natural language. Among these approaches, CLIP showed that large-scale image-text pre-training can learn a shared embedding space in which visual and textual representations are semantically aligned~\cite{radford2021learning}.

Subsequent studies have explored prompt-based adaptation to transfer pre-trained vision-language models to downstream tasks. In particular, conditional prompt learning approaches such as CoCoOp~\cite{zhou2022conditional} demonstrated that textual prompts can be optimized to better reflect task-specific semantic concepts. This is particularly relevant to expression recognition, where each expression class can be associated with a linguistic description that provides additional semantic priors.

Recent studies have applied CLIP-based representations to expression recognition~\cite{zhou2025emotion}. In the ABAW context, CLIP has also been explored as a visual feature extractor and in conjunction with rule-based emotion prompts~\cite{cabacasmaso2025enhancingfacialexpressionrecognition}. These studies suggest that vision-language representations can provide useful semantic guidance for expression recognition under challenging in-the-wild conditions.

\subsection{Temporal Modeling}

Temporal modeling has become increasingly important in expression analysis, since facial expressions evolve over time through subtle movements and intensity changes. Static expression recognition methods can capture discriminative spatial patterns from individual images, but often fail to reflect the process of expression formation and transition. This limitation has motivated the development of video-based and dynamic expression recognition approaches.

Recent studies have increasingly adopted transformer-based architectures to capture both spatial and temporal information in dynamic expression recognition. Former-DFER~\cite{zhao2021former} introduced separate modules for learning spatial and temporal representations, while Spatio-Temporal Transformer (STT) jointly encoded spatial and temporal cues within a unified transformer framework~\cite{ma2022spatio}. NR-DFERNet~\cite{li2022nrdfernet} addressed the influence of noisy or less informative frames, highlighting the importance of both temporal continuity and frame quality in robust temporal modeling.

Beyond transformer-based approaches, convolution-based methods have also been explored for temporal sequence modeling. In particular, Temporal Convolutional Networks (TCNs) model temporal dependencies efficiently through convolutional operations with enlarged receptive fields and have shown strong performance in generic sequence modeling tasks~\cite{bai2018empirical}. TCNs have also been applied to video-based temporal understanding problems, further suggesting their applicability beyond recurrent modeling~\cite{lea2016temporal}.

\subsection{Multimodal Fusion}

Multimodal learning has been widely studied for learning shared representations from multiple information sources~\cite{ngiam2011multimodal}. In emotion recognition, this perspective is particularly important because affective cues are distributed across multiple modalities, including facial appearance and speech.

Among these modalities, visual and audio information are particularly useful for emotion recognition. Facial expressions provide explicit visual evidence of affective changes, while speech conveys additional emotional cues through tone, prosody, and temporal variation. To obtain informative audio representations, recent studies have explored self-supervised speech representation learning. In particular, wav2vec 2.0~\cite{baevski2020wav2vec20frameworkselfsupervised} introduced an effective framework for learning contextualized speech representations from raw audio, making it useful for downstream audio-based affective analysis.

The interaction between visual and audio modalities is a key factor in multimodal emotion recognition. Recent studies have explored transformer-based fusion methods to model cross-modal dependencies more effectively than simple feature concatenation~\cite{tsai2019multimodal,tran2022pre,venkatraman2024multimodal}. By allowing one modality to attend to another, these methods can capture complementary affective cues distributed across facial and vocal streams.

Several ABAW approaches have explored multimodal frameworks that integrate multiple modalities, including visual, audio, and in some cases linguistic information, for affective behavior analysis~\cite{zhang2022continuous,zhang2024affective,zhou2023leveraging}. However, effectively integrating robust audio representation learning, visual temporal modeling, and cross-modal interaction remains an open challenge in expression recognition. This motivates a framework that integrates CLIP-based visual representation learning, wav2vec 2.0-based audio representation learning, visual temporal modeling, and transformer-based multimodal fusion for robust expression recognition. 
\section{Approach}
\label{sec:method}

In this section, we describe the proposed multimodal emotion recognition framework for the EXPR task. Our model consists of five components: 
(1) unimodal feature extraction using pre-trained backbones, 
(2) temporal visual modeling and audio feature adaptation, 
(3) bi-directional cross-attention fusion, 
(4) temporal pooling and emotion classification, and 
(5) label-based prompt supervision via contrastive learning.

\begin{figure*}[t]
    \centering
    \includegraphics[width=\linewidth]{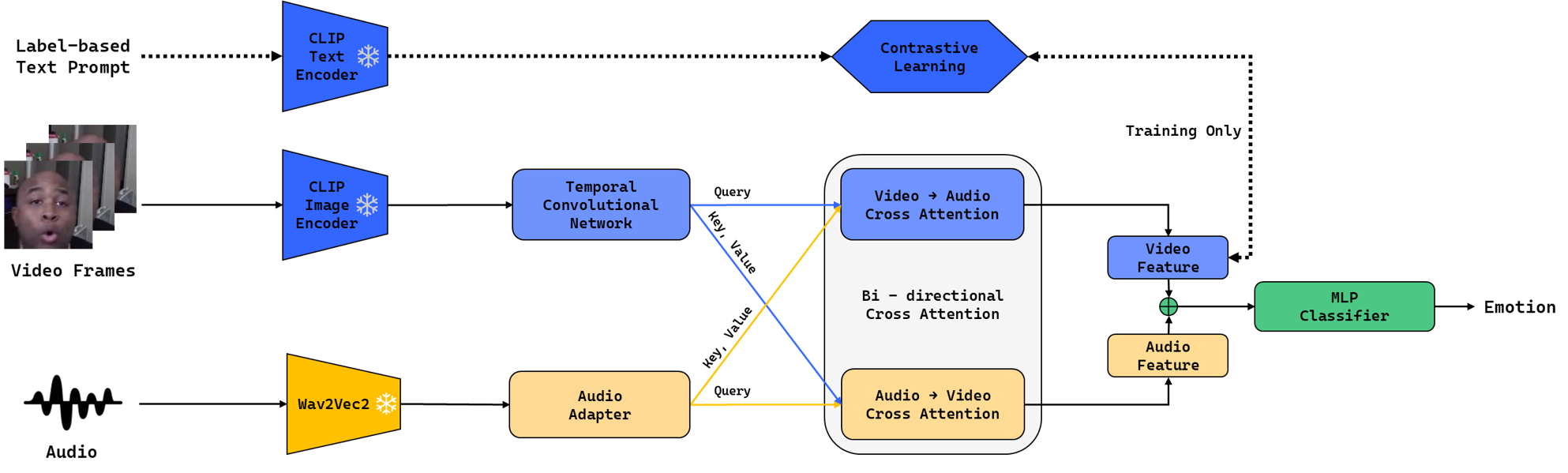}
    \caption{Overall architecture of the proposed multimodal framework for the EXPR task. The model extracts visual and audio features, enhances them through temporal modeling and bi-directional cross-attention fusion, and predicts expression categories with auxiliary label-based prompt supervision during training. Video frames are encoded by a frozen CLIP image encoder followed by a temporal convolutional network, while audio signals are processed by a frozen Wav2Vec2 backbone and an audio adapter. The resulting representations are fused by bi-directional cross-attention, temporally pooled, and fed into an MLP classifier. During training, class-specific text prompts derived from expression labels are encoded by the CLIP text encoder and aligned with projected visual features via a contrastive objective.}
    \label{fig:model_architecture}
\end{figure*}

\subsection{Feature Extraction and Text Prompt Construction}

To obtain high-level semantic representations from raw visual and audio inputs, we employ large-scale pre-trained models as frozen backbone networks.

\begin{itemize}

\item \textbf{Visual Stream.}
For the visual modality, we use the CLIP ViT-B/32 image encoder.
Given an input frame sequence, each frame is independently encoded to produce a visual feature sequence $X_v \in \mathbb{R}^{T_v \times D_v}$, where $D_v = 512$.

\item \textbf{Audio Stream.}
For the audio modality, an audio segment synchronized with the input video window is resampled to 16 kHz and processed by the frozen Wav2Vec 2.0 Base model, producing an audio feature sequence $X_a \in \mathbb{R}^{T_a \times D_a}$, where $D_a = 768$.

\item \textbf{Text Prompts.}
To provide semantic supervision for emotion recognition, we generate a text prompt corresponding to the target emotion label using the template
\textit{``A face expressing [Emotion]''}.
Each prompt $p_j$ is encoded by the frozen CLIP text encoder to produce a text embedding $\tilde{t}_j = E_{\text{text}}(p_j)$.

\end{itemize}

To improve training stability and reduce overfitting on the downstream benchmark, both the CLIP and Wav2Vec 2.0 backbones are kept frozen during training.

\subsection{Temporal Modeling and Audio Feature Adaptation}

Facial expressions evolve over time, and frame-level representations alone may not sufficiently capture temporal dynamics. Therefore, we further refine the unimodal features before multimodal fusion.

\begin{itemize}

\item \textbf{Visual TCN.}
To model temporal dependencies in the visual stream, we feed the frame-level visual features $X_v$ into a Temporal Convolutional Network (TCN) composed of six stacked temporal blocks with dilated causal convolutions. By progressively increasing the dilation factor across layers, the TCN expands the temporal receptive field and captures both short- and long-range facial dynamics. The final refined visual representation is denoted as $F_v \in \mathbb{R}^{T_v \times 512}$.

\item \textbf{Audio Adapter.}
The audio features $X_a$ are projected into the same embedding dimension as the visual features using an audio adapter, which consists of a linear projection followed by Layer Normalization, ReLU activation, and Dropout. This produces the adapted audio representation $F_a \in \mathbb{R}^{T_a \times 512}$. Matching the embedding dimensions facilitates effective cross-modal interaction in the subsequent fusion stage.

\end{itemize}

\subsection{Bi-directional Cross-Attention Fusion}

To effectively integrate visual and audio information, we employ a bi-directional cross-attention module with eight attention heads. 
Cross-modal attention is performed in both directions to enable mutual interaction between the two modalities.

First, the visual features $F_v$ are used as queries, while the audio features $F_a$ are used as keys and values, yielding an audio-enhanced visual representation:
\begin{equation}
H_{V2A} = \mathrm{LN}\big(F_v + \mathrm{MHA}(F_v, F_a, F_a)\big).
\end{equation}

Second, the audio features $F_a$ are used as queries, while the visual features $F_v$ are used as keys and values, yielding a visually enhanced audio representation:
\begin{equation}
H_{A2V} = \mathrm{LN}\big(F_a + \mathrm{MHA}(F_a, F_v, F_v)\big).
\end{equation}

\subsection{Temporal Pooling and Emotion Classification}

After cross-attention fusion, the enhanced feature sequences are aggregated using mean pooling along the temporal dimension. The pooled representations are concatenated to form the multimodal feature vector $z \in \mathbb{R}^{1024}$. This representation is fed into a multi-layer perceptron (MLP) classifier consisting of three linear layers with ReLU activations and Dropout, producing the final logits for the eight emotion classes.

\subsection{Label-Based Prompt Supervision via Contrastive Learning}

In addition to the cross-entropy classification loss $\mathcal{L}_{cls}$, we use class-specific text prompts derived from expression labels as an auxiliary supervision signal via a contrastive objective, encouraging label-aware visual representations.

The temporally pooled visual representation after cross-modal fusion is projected into the CLIP embedding space and $\ell_2$-normalized to obtain $v$. Similarly, the text embeddings are normalized as $t_j$. Using scaled cosine similarity, we compute a bidirectional contrastive loss $\mathcal{L}_{con}$ between matched video-text pairs within a mini-batch.

Finally, the overall training objective combines the classification loss and the contrastive loss:

\begin{equation}
\mathcal{L} = \mathcal{L}_{cls} + \lambda \mathcal{L}_{con},
\end{equation}

where we set $\lambda = 0.1$ in our implementation.   
\section{Experiments}
\label{sec:experiment}

In this section, we evaluate the proposed multimodal framework on the EXPR task of the ABAW 10th Challenge. We first describe the implementation details and evaluation metrics, and then examine how different temporal window sizes affect recognition performance, with comparisons to the official baseline.

\subsection{Implementation and Evaluation Setup}
The proposed model was implemented in PyTorch and trained on an NVIDIA RTX A5000 GPU. We used frozen CLIP ViT-B/32 and Wav2Vec 2.0 backbones for the visual and audio streams, respectively. All experiments were conducted with a fixed random seed of 42 for reproducibility.

For optimization, we used AdamW with a learning rate of $1\times10^{-5}$ and a cosine annealing scheduler. The model was trained for 30 epochs with a batch size of 64 and gradient accumulation. To alleviate class imbalance in the EXPR dataset, we employed a class-weighted cross-entropy loss for expression classification. In addition, text-guided contrastive learning was used during training as an auxiliary supervision objective.

\subsection{Evaluation Metrics}
Following the official protocol of the ABAW 10th EXPR Recognition Challenge, we report the following evaluation metric:
\begin{itemize}
    \item \textbf{Macro F1-score:} The official primary metric of the challenge, computed by averaging the F1-scores of all eight expression classes equally.
\end{itemize}

\subsection{Results on the Validation Set}
Table~\ref{table:window_size_results} presents the validation performance of the proposed method under different temporal window sizes, together with the official baseline. We report results for four settings, namely 10, 15, 30, and 60 frames, to analyze the effect of temporal context.

\begin{table}[ht]
\centering
\renewcommand{\arraystretch}{1.25}
\caption{Performance comparison on the EXPR challenge validation set.}
\label{table:window_size_results}
\begin{tabular}{cclc}
\hline
\multicolumn{1}{c}{\textbf{Challenge}} & 
\multicolumn{1}{c}{\textbf{Metric}} & 
\multicolumn{1}{c}{\textbf{Method}} & 
\multicolumn{1}{c}{\textbf{Result}} \\ \hline
\multirow{5}{*}{EXPR} & \multirow{5}{*}{Macro F1} & Baseline & 0.2500 \\
 &  & Ours (10 frames) & 0.3265 \\
 &  & Ours (15 frames) & 0.3324 \\
 &  & Ours (30 frames) & \textbf{0.3670} \\
 &  & Ours (60 frames) & 0.3599 \\ \hline
\end{tabular}
\end{table}

As shown in Table~\ref{table:window_size_results}, the proposed multimodal framework outperforms the official baseline across all temporal window sizes. Among the evaluated settings, the 30-frame configuration achieves the best performance, with a Macro F1-score of 0.3670. The results indicate that incorporating temporal context is beneficial for expression recognition, while increasing the temporal window beyond a certain length does not necessarily lead to further improvement.

Overall, these results show that the proposed framework provides consistent performance gains over the baseline on the validation set.

\subsection{Results on Test Set}
As shown in Table~\ref{table:test_set_results}, the 60-frame configuration achieved the best performance on the test set among the evaluated settings. It also outperformed the baseline score of 0.25.

\begin{table}[ht]
\centering
\renewcommand{\arraystretch}{1.25}
\caption{Performance comparison on the EXPR challenge test set.}
\label{table:test_set_results}
\begin{tabular}{cclc}
\hline
\multicolumn{1}{c}{\textbf{Challenge}} & 
\multicolumn{1}{c}{\textbf{Metric}} & 
\multicolumn{1}{c}{\textbf{Method}} & 
\multicolumn{1}{c}{\textbf{Result}} \\ \hline
\multirow{5}{*}{EXPR} & \multirow{5}{*}{Macro F1} & Baseline & 0.25 \\
 &  & Ours (10 frames) & 0.28 \\
 &  & Ours (15 frames) & 0.30 \\
 &  & Ours (30 frames) & 0.31 \\
 &  & Ours (60 frames) & \textbf{0.32} \\ \hline
\end{tabular}
\end{table}
\section{Conclusion}
\label{sec:conclusion}

In this paper, we presented a multimodal emotion recognition framework for the EXPR task of the ABAW 10th Challenge. The proposed model combines CLIP-based visual representations and Wav2Vec 2.0-based audio representations within a unified architecture. To better model temporal and cross-modal information, we introduced a Temporal Convolutional Network (TCN) for visual temporal modeling and a bi-directional cross-attention module for multimodal fusion. In addition, a text-guided contrastive objective was incorporated to encourage semantically aligned visual representations.

Experimental results on the validation set demonstrated that the proposed framework outperforms the official baseline in terms of Macro F1-score. In particular, the 60-frame setting achieved the best performance, indicating that incorporating a broader temporal context is beneficial for robust expression recognition in unconstrained environments.

For future work, we plan to explore more effective temporal modeling strategies and stronger multimodal fusion mechanisms. We also aim to investigate the use of additional modalities and more efficient architectures for robust emotion recognition in real-world scenarios.
{
    \small
    \bibliographystyle{ieeenat_fullname}
    \bibliography{main}
}

\end{document}